\documentclass{article}
\usepackage{fix-cm}
\usepackage{mlspconf}
\usepackage{adjustbox}
\usepackage{cite}
\usepackage{amsmath,amssymb,amsfonts}
\usepackage{placeins}
\usepackage{algorithmic}
\usepackage{textcomp}
\usepackage{bm}
\usepackage{graphicx}
\graphicspath{ {./Figures/} }
\usepackage{subfig}
\usepackage{xcolor}

\usepackage{authblk}
\def\BibTeX{{\rm B\kern-.05em{\sc i\kern-.025em b}\kern-.08em
    T\kern-.1667em\lower.7ex\hbox{E}\kern-.125emX}}

\usepackage{tabularx,ragged2e,booktabs,caption}
\newcolumntype{C}[1]{>{\Centering}m{#1}}

\renewcommand{\tabcolsep}{2pt}

\usepackage[commentmarkup=uwave]{changes}
\definechangesauthor[]{FC}

\copyrightnotice{978-1-7281-6338-3/21/\$31.00 {\copyright}2021 IEEE}

\toappear{2021 IEEE International Workshop on Machine Learning for Signal Processing, Oct. 25--28, 2021, Gold Coast, Australia}

\title{Uncertain Bayesian Networks: Learning from Incomplete Data}

\name{Conrad D. Hougen$^1$, Lance M. Kaplan$^2$, Federico Cerutti$^3$, Alfred O. Hero$^1$ 
\thanks{Primary Author Email: chougen@umich.edu. This research was partially supported by the Army Research Office under contract W911NF-15-1-0479 and partially supported by the U.S. Army Research Laboratory and the U.K. Ministry of Defence under agreement W911NF-16-3-0001. The views and conclusions contained in this document are those of the authors and should not be interpreted as representing the official policies, either expressed or implied, of the U.S. Army Research Laboratory, the U.S. Government, the U.K. Ministry of Defence or the U.K. Government. The U.S. and U.K. Governments are authorized to reproduce and distribute reprints for Government purposes notwithstanding any copyright notation hereon.}}
\address{$^1$Univ. of Michigan, $^2$US Army DEVCOM ARL, $^3$Univ. of Brescia}

\begin{document}
\raggedbottom
\ninept

\maketitle
\begin{abstract}
When the historical data are limited, the conditional probabilities associated with the nodes of Bayesian networks are uncertain and can be empirically estimated. Second order estimation methods provide a framework for both estimating the probabilities and quantifying the uncertainty in these estimates. We refer to these cases as uncertain or second-order Bayesian networks. When such data are complete, i.e., all variable values are observed for each instantiation, the conditional probabilities are known to be Dirichlet-distributed. This paper improves the current state-of-the-art approaches for handling uncertain Bayesian networks by enabling them to learn distributions for their parameters, i.e., conditional probabilities, with incomplete data. We extensively evaluate various methods to learn the posterior of the parameters through the desired and empirically derived strength of confidence bounds for various queries. 

\end{abstract}


\section{Introduction}

Bayesian networks (Section \ref{bg}), or \emph{BNs}, provide a well-established paradigm in the probabilistic machine learning literature. Traditionally, the conditional probabilities of a BN are known, either from historical data or from subject matter experts. However, when the historical data or prior knowledge are limited, the conditional probabilities may not be known precisely. From a Bayesian perspective, the conditional probabilities are known within a posterior distribution representing epistemic uncertainty due to limited training examples. When historical data are complete, i.e., all variable values are observed for each instantiation, the conditional probabilities are known to be Dirichlet-distributed. 

Several prior works exist in the literature for handling uncertain Bayesian inference. For instance, MeanVAR \cite{VanAllen} combines the ``delta-rule'' with bucket elimination to approximate the variances of the conditional probabilities and avoids generating large numbers of Monte Carlo samples. Despite this, MeanVAR is not scalable to large networks when querying many variables, as each variable is inferred separately. Other methods are based on inferencing belief networks using imprecise probabilities, including \cite{shenoy}, which utilizes Dempster-Shafer theory, and \cite{zaffalon}, which utilizes creedal sets. However, these methods are not rooted in probability theory. In fact, it is shown in \cite{kaplan-ivanovska} that the methods developed in \cite{shenoy} and \cite{zaffalon} underestimate or overestimate the uncertainties, respectively, when we have distributional knowledge of the conditional probabilities. Recently, the technique proposed in \cite{kaplan-ivanovska} extended Judea Pearl's belief propagation for second-order probabilistic inference but is only exact for polytree networks, not arbitrary DAGs. A survey of some of the common techniques for Bayesian networks is presented in \cite{rohmer}. Going beyond BNs, the method in \cite{cerutti} enables second-order probabilistic inferencing over any arbitrary directed acyclic graph.

In this paper we go beyond the inference problem by improving the current state-of-the-art approaches for learning the parameters of uncertain (second-order) Bayesian networks with incomplete data. Here, the traditional approaches exploit Monte Carlo sampling, with the inevitable computational overhead that comes with them. We, instead, aim at approximating the resulting distributions as Dirichlet, while being as accurate as possible at assessing their (co)variances. 

This paper (Section \ref{meth}) therefore provides the first evaluations of algorithms capturing both the mean and covariance of the uncertain conditional probabilities for later second-order inferencing (see\cite{kaplan-ivanovska,cerutti}). Techniques based upon the method of moments, Gaussian approximation and the Fisher information matrix are considered.
This paper extends \cite{kaplan}, which was limited to the special case of a two-node, binary BN, to arbitrary Bayesian networks for categorical variables with finite domain sizes. Moreover, while such techniques are not new in the Bayesian community, in this paper we rigorously assess their ability to provide accurate estimates of the posterior distributions' covariances.



We indeed extensively test (Section \ref{sim}) second-order queries using the learned parameters via the the three learning methods in terms of (1) accuracy of the computation of the covariance matrix, which lies at the heart of uncertain Bayesian networks, through consistency of the desired and empirically derived strength of confidence bounds; and (2) their scalability. Improving the latter is one of the main areas of future work (Section \ref{conc}), especially when it comes to the approach of using the inverse of the Fisher information matrix as the approximate covariance.

\section{Background}\label{bg}
\subsection{Bayesian Networks}
\par
A Bayesian network is a directed acyclic graph where each node is a random variable and links represent dependencies \cite{nielsen}: each child variable $A$ with parents $B_1,...,B_k$ has an associated conditional probability table $p(A=a|B_1=b_1,...,B_k=b_k)$, and a variable, $X$, with no ancestors, is associated with an unconditional probability table $p(X)$. In this paper we assume that each variable has a finite set of mutually exclusive states.

A given graph with fixed structure---i.e., the set of nodes and edges---represents a family of Bayesian networks parameterized by the probabilities $\bm{\theta}$, which we interchangeably refer to as the ``parameters'' or ``conditionals.'' Each parameter corresponds to a particular entry in the conditional or unconditional probability tables of the BN.
The joint probability of variables $X_1,...,X_n$ is
\begin{equation}\label{eqn:jointprob}
    p(x_1,...,x_n; \bm{\theta})=\prod_{i=1}^{n} \theta_{x_i|pa_i}.
\end{equation}
where $x_i\in \mathbb{X}_i$ is the value that $X_i$ is assigned, and $pa_i\in \mathbb{PA}_i$ represents a specific assignment of the parents of variable $X_i$. $\mathbb{X}_i$ and $\mathbb{PA}_i$ denote the sets of possible values for $X_i$ and the parents of $X_i$, respectively.  It is worth noticing that 
\begin{equation}\label{eqn:secondconstraint}
    \sum_{x_i\in \mathbb{X}_i} \theta_{x_i|pa_i} = 1, \forall pa_i\in \mathbb{PA}_i, \forall i.
\end{equation}

\subsection{Sum-Product Networks}\label{spns}
In this paper, we consider only the case of discrete-valued, finite domain variables. This enables the learning methods to leverage sum-product networks (SPNs).  As discussed in \cite{Zhao}, SPNs and BNs are equally expressive when representing the joint probability distribution of discrete variables. In fact, any BN with discrete-valued variables of finite domains can be transformed into an SPN via the well-known technique pioneered by Darwiche based on variable elimination \cite{darwiche}. Note that a single BN may be associated with multiple different SPNs, based on the order of variable elimination performed. This means it might be possible to find the most computationally efficient SPN for learning uncertain BNs. This paper assumes a specified order for variable elimination, and exploitation of the most efficient SPN is beyond the scope of this paper.


An SPN consists of a rooted DAG of interior operator nodes (e.g. sum or product operations) and leaf nodes which are associated with indicator variables. In order to convert a BN into an SPN, the basic idea is to introduce indicator variables $\lambda_{x_i};\ i=1,\ldots,n$ such that
\begin{equation}
    \lambda_{x_i} = \begin{cases}
      1, & \text{if}\ X_i = x_i \\
      0, & \text{otherwise}
    \end{cases}.
\end{equation}
When combined with the BN parameter variables $\bm{\theta}$, we may write down a canonical polynomial for any particular BN. Given the canonical polynomial, it is relatively straightforward to form an associated SPN graphical model, by means of variable elimination. The specific SPN structure is not unique and will depend on the order in which variables are eliminated. We refer the reader to \cite{darwiche} for more details.

The indicator variables of the leaf nodes in an SPN represent the assignment states of variables from the original BN from which the SPN was derived. By setting the indicator nodes of the SPN to either 1 or 0, we can efficiently compute $p(\bm{e})$, where $\bm{e}$ represents the evidence, i.e. the values of the observed variables. In order to accomplish this, we set the values of the leaf indicators to 0 if the associated conditional probabilities are inconsistent with the evidence. A forward pass from the leaf nodes to the root of the SPN performs marginalization of eqn.~(\ref{eqn:jointprob}) over all variables which are unobserved:
\begin{equation}
    p(\bm{e})=\sum_{X_i\notin \bm{E}} \prod_{i=1}^n \theta_{x_i|pa_{i}}.
\end{equation}
An SPN readily performs this operation during a single forward pass.

Additionally, a backward pass through an SPN can be used to efficiently compute partial derivatives of the evidence likelihood with respect to each of the BN parameters, i.e. partial derivatives of the form
$$\frac{\partial p(\mathbf{e};\bm{\theta})}{\partial \theta_{x_k|{\mathbf{pa}_k}}}.$$
These partial derivatives are beneficial for computing joint probabilities $p(x_k,\bm{e})$, for each of the variables $x_k$. These values are output at the leaf nodes of the SPN:

\begin{equation}
    p(x_k,\bm{e})=\sum_{pa_{k}\in \mathbb{PA}_k} \frac{\partial p(\mathbf{e};\bm{\theta})}{\partial \theta_{x_k|{\mathbf{pa}_k}}}\theta_{x_k|\mathbf{pa}_k}.
\end{equation}

\section{Methods for Learning with Incomplete Data}\label{meth}

In the work, we consider three methods to approximate the posterior distributions of the parameters.  These methods are well known in the Bayesian learning literature.  However, to the best of our knowledge they have only been employed to determine the mean probability values for Bayesian network queries.  Here, we are incorporating the learned distributions for second-order inferences via \cite{cerutti} to evaluate the accuracy of the returned distributions for the probabilities of the queried values. The first method incorporates moment matching in a sequential fashion. The last two use expectation-maximization (EM) to compute the means and either Gaussian approximation or the Fisher information matrix to approximate the covariances. 

\subsection{Method 1: Online Bayesian Moment Matching}

Given only partially-observed training data, estimates of the parameters $\bm{\theta}$ are generated by various second-order Bayesian learning algorithms. One of the most prominent approaches is online Bayesian Moment Matching (BMM) \cite{rashwan}. 

The BMM method approximates the posterior distribution as a product of Dirichlet random variables, presuming that the group of conditional probabilities are statistically independent. It is known that the ground-truth, i.e. latent distribution, of $\bm{\theta}$ is Dirichlet, so we begin with a prior that is a product of Dirichlets with respect to the weights of each sum node in the sum product network. From \cite{rashwan}, we have
\begin{equation} \label{dirprod}
\begin{split}
f^{(0)}(\bm{\theta}) &=   \prod_{i\in sumNodes} Dir(\bm{\theta}_i;\bm{\alpha}^{(0)}_i)\\
& = \prod_{\bm{\theta}_{X_i,pa_i} \in \mathcal{F}} Dir(\bm{\theta}_{X_i|pa_i};\bm{\alpha}^{(0)}_{X_i,pa_i})
\end{split}
\end{equation}
as the prior distribution, where $\bm{\alpha}_{X_i,pa_i}^{(t)}$ is the vector of shape parameters associated with each Dirichlet distribution. In eqn. (\ref{dirprod}), $\mathcal{F}$ represents the set of families of parameters where each parameter family sums to $1$, i.e.
\begin{equation}
    \mathcal{F} = \{\{\theta_{x_i|pa_i}:x_i \in \mathbb{X}_i\}: i \in [n], pa_i \in \mathbb{PA}_i \}.
\end{equation}

We initialize the Dirichlet shape parameters as $\bm{\alpha}_i^{(0)}=(1,1,\ldots,1)$ for the natural uniform prior, then 
the posterior distribution after $T$ training instances is 
\begin{equation}\label{eqn:posterior_T}
f(\bm{\theta}|\{\mathbf{e}_{t}\}_{t=1}^T) \propto \prod_{t=1}^T \left ( \sum_{(x_i,\mathbf{pa}_i)\sim\mathbf{e}_t} \theta_{x_i|\mathbf{pa}_i} \frac{\partial p(\mathbf{e};\bm{\theta})}{\partial \theta_{x_i|{\mathbf{pa}_i}}} \right). 
\end{equation}
The evaluation of a given SPN consists of alternating sums and products, thereby resulting in a polynomial representation of eqn. (\ref{eqn:posterior_T}) with respect to the parameters, which means that the posterior becomes a mixture of products of Dirichlets. While the mixture of Dirichlet products admits a closed form expression for its posterior distribution, unfortunately it is also computationally intractable since the number of mixture components is exponential in the number of sum nodes in the SPN \cite{rashwan}. BMM solves this problem by presuming the posterior after incorporation of $t$ instantiations is a product of Dirichlets, i.e., $f^{(t)}(\bm{\theta}) \approx f(\bm{\theta}|\{\mathbf{e}_{t'}\}_{t'=1}^t) $ where
\begin{equation}
  f^{(t)}(\bm{\theta}) = \prod_{i=1}^n \prod_{\mathbf{pa}_i\in\mathbb{PA}_i} Dir(\theta_{X_i|\mathbf{pa}_i};\bm{\alpha}^{(t)}_{X_i|\mathbf{pa}_i})
\end{equation}
and fitting a product of Dirichlets to the posterior after $t+1$ instantiations, i.e.,
\begin{equation}
   f^{(t+1)}(\bm{\theta}) =\underbrace{\left(\sum_{x_i \sim\mathbf{e}_{t+1}}\sum_{\mathbf{pa}_i\sim\mathbf{e}_{t+1}} \theta_{x_i|\mathbf{pa}_i} \frac{\partial p(\mathbf{e};\bm{\theta})}{\partial \theta_{x_i|{\mathbf{pa}_i}}}\right)}_{p(\bm{e}_{t+1};\bm{\theta})} f^{(t)}(\bm{\theta}),
\end{equation}
via the method of moments. 


Specifically, the first and second order moments of the parameters are
\begin{equation}
m^{(t)}_{x_i|\mathbf{pa}_i} = E[\theta_{x_i|\mathbf{pa}_i}] = \frac{Z[1]}{Z[0]}, 
\end{equation}
\begin{equation}
v^{(t)}_{x_i|\mathbf{pa}_i} = E[\theta^2_{x_i|\mathbf{pa}_i}] = \frac{Z[2]}{Z[0]},
\end{equation}
where $Z[k]$ is given by
\begin{equation}
    Z[k;\theta_{x_i|\mathbf{pa}_i}] = \int  \theta^k_{x_i|\mathbf{pa}_i}p(\mathbf{e}_{t+1};\bm{\theta}) f^{(t)}(\bm{\theta}) d\mathbf{\theta}
\end{equation}
and can easily be computed in closed form by leveraging the properties of Dirichlet distributions.

The posterior after $t+1$ instantiations can be approximated as a product of Dirichlets by using the values of $m^{(t)}_{x_i|\mathbf{pa}_i}$ and $v^{(t)}_{x_i|\mathbf{pa}_i}$ to determine $\alpha^{(t)}_{x_i|\mathbf{pa}_i}$:
\begin{equation}
    \alpha^{(t+1)}_{x_i|\mathbf{pa}_i} = m^{(t)}_{x_i|\mathbf{pa}_i} S^{(t+1)}_{X_i|\mathbf{pa}_i}
\end{equation}

\begin{equation}
    S^{(t+1)}_{X_i|\mathbf{pa}_i}= \frac{\sum_{x_i\in\mathbb{X}_i} m^{(t)}_{x_i|\mathbf{pa}_i} (1-m^{(t)}_{x_i|\mathbf{pa}_i}) (m^{(t)}_{x_i|\mathbf{pa}_i}-v^{(t)}_{x_i|\mathbf{pa}_i})}{\sum_{x_i\in\mathbb{X}_i} m^{(t)}_{x_i|\mathbf{pa}_i} (1-m^{(t)}_{x_i|\mathbf{pa}_i}) (v^{(t)}_{x_i|\mathbf{pa}_i}-(m^{(t)}_{x_i|\mathbf{pa}_i})^2)}.
\end{equation}
Note that this moment matching technique exactly matches the means and sets the Dirichlet strength $S^{(t+1)}_{X_i|\mathbf{pa}_i}$ to minimize the mean square differences between the variances. 


Notice that in eqn. (\ref{dirprod}), we implicitly make the assumption that the parameters $\bm{\theta}$ are independent, which is why the probability of $\bm{\theta}$ appears as a product. This assumption affects the performance of the moment matching approach for certain pathological cases where the available data do not allow for the justification of such an assumption.

\subsection{Expectation Maximization for Methods~2 and~3}




Our objective is to accurately estimate both the posterior mean and covariance when training uncertain Bayesian networks with partial data.
When complete observations of the network variables are available at each training instance, the free variables are statistically independent. This implies that the covariance matrix of the parameters is block diagonal. One can verify this by observing the form of the posterior, which can be analytically computed in cases where complete observations are present, as seen in \cite{kaplan}. Here, however, we are dealing with incomplete observations, where statistical independence between the network parameters may or may not be a good assumption. The independence assumption may be reasonable when the incomplete training data contain enough samples to enable efficient estimation of each parameter. Otherwise, the assumption may be poor. 





In this paper we base our proposal on an EM algorithm. Indeed, let us note 
that, in general, given the observed training data, a straightforward approach would be to estimate the parameters as the maximum a posteriori (MAP) estimate
\begin{equation} \label{eqn:map}
    \bm{\theta} = \text{argmax}_{\bm{\theta}} \log((P(X|\bm{\theta})f(\bm{\theta}))
\end{equation}
where $f(\bm{\theta})$ is a prior distribution for the parameters. We would normally model this prior as a Dirichlet distribution. In the case where we have complete data observed, we can easily compute (\ref{eqn:map}) since the logarithm of the product decomposes into a sum of logarithms. Unfortunately, with incomplete data, this is not possible since latent variables will prevent the simple decomposition.

The EM algorithm helps us by providing simple iterative decompositions at each step. The algorithm proceeds as a two step process, starting with the expectation step
\begin{equation}
    Q(\bm{\theta};\bm{\theta}^{(t)})=\\
    \sum_{X_\ell\in \mathbb{X}_\ell} \log(P(X_o,X_\ell;\bm{\theta})f(\bm{\theta}))P(X_\ell|X_o;\bm{\theta}^{(t)})
\end{equation}
where $X_\ell$ are the unobserved latent variables and $X_o$ are the observed variables. The maximization step, which updates the estimated parameters, is given by
\begin{equation}
    \bm{\theta}^{(t+1)}=\text{argmax}_{\bm{\theta}} Q(\bm{\theta};\bm{\theta}^{(t)}).
\end{equation}

The EM algorithm starts with an initial parameter estimate and then iterates over the E and M steps until convergence. The result is an estimate of the mean values of the parameters, but note that we must go further to estimate the confidence levels of these values. Namely, we next estimate the covariance of the parameters. In this paper, we achieve this covariance estimation step through two methods: using a Gaussian approximation of the parameters to estimate the covariance matrix (EM-GA); and inverting the Fisher information matrix as an estimate of the parameter covariance matrix (EM-Fisher).

\subsection{Method~2: EM-GA}
In the second method, we assume a Gaussian approximation of the parameters in order to estimate the covariance matrix. In the Gaussian approximation, we approximate the covariance matrix as
\begin{equation}\label{eqn:EM-GA_inversion}
    R = D^T(DHD^T)^{-1}D,
\end{equation}
where
\begin{equation}
H \approx J_0 + \sum_t \frac{1}{p^2(\mathbf{e}_t)} \nabla_\mathbf{\theta} p(\mathbf{e}_t) \nabla^T_\mathbf{\theta} p(\mathbf{e}_t),
\end{equation}
and $J_0 = K\mbox{diag}(\frac{1}{\bm{\theta}})$ where $K$ is a diagonal matrix with values corresponding to the cardinality of the corresponding child domain $|\mathbb{X}_i|$. The matrix $D$ is a matrix which transforms the partial derivatives over $\bm{\theta}$ into the full derivatives for each of the free parameters. For instance, $\theta_{X_i|\mathbf{pa}_i}$ is associated to $k=|\mathbb{X}_i|$ parameters for which $k-1$ are free since they are constrained to sum up to one. In other words for a fixed assignment of $\mathbf{pa}_i$, knowing $k-1$ of the parameters, $\theta_{x_{i,1}|\mathbf{pa}_i},\theta_{x_{i,2}|\mathbf{pa}_i},...,\theta_{x_{i,k-1}|\mathbf{pa}_i}$ uniquely determines the value of $\theta_{x_{i,k}|\mathbf{pa}_i}$. Therefore,
\begin{equation}
    \dfrac{dp(\bm{e};\bm{\theta})}{d\theta_{x_{i,j}|\mathbf{pa}_i}} = \frac{\partial p(\mathbf{e};\bm{\theta})}{\partial \theta_{x_{i,j}|{\mathbf{pa}_i}}} - \frac{\partial p(\mathbf{e};\bm{\theta})}{\partial \theta_{x_{i,k}|{\mathbf{pa}_i}}},  \forall j\in\{1,...,k-1\}.
\end{equation}

This method takes advantage of SPNs, as discussed in Section \ref{spns}, in order to efficiently compute the partial derivatives of the evidence likelihood. These partial derivatives, when joined into a vector, form the gradient of the likelihood, and the outer product of these likelihood gradients gives the Hessian matrix $H$.

\subsection{Method~3: EM-Fisher}
In the third method, we invert the Fisher information matrix (FIM) as an estimate of the parameter covariance matrix. The posterior distribution of the parameters is asymptotically normal with covariance given by the inverse FIM, per the Bernstein-von Mises theorem. For incomplete training data, there is a known form for the FIM, i.e.
\begin{equation}\label{eqn:EM-Fisher}
    J = J_0 + \sum_t \sum_{\mathbf{e}' \in \mathbb{E}_t} \frac{1}{p^2(\mathbf{e}')}\nabla_\theta p(\mathbf{e}')
    \nabla_\theta^T p(\mathbf{e}').
\end{equation}
See \cite{kaplan} for a derivation. The parameter covariance is then given by 
\begin{equation}\label{eqn:EM-Fisher_inversion}
    R = D^T(DJD^T)^{-1}D.
\end{equation}
Again, since the FIM is derived from the Hessian, i.e. the outer product of the likelihood gradients, we take advantage of SPNs to efficiently compute the appropriate partial derivatives.


\section{Experimental Analysis}\label{sim}

\subsection{Experimental Approach}
For quantifying the accuracy of the distributional estimates of the Bayesian network parameters, we employed the desired confidence bound divergence (DeCBoD) as in \cite{kaplan}, for which we describe the experimental setup here. After fixing a specific BN graph structure, we randomly generated $N=1000$ complete assignments of the BN conditional probabilities. Given the BN parameters, a unique set of $T$ observations were then drawn for each of the $N$ Bayesian networks, each set forming the training data for that particular BN. The ground truth conditional probabilities for each instantiation of the BN were recorded. This allowed us to construct confidence intervals of significance level $\gamma$ around each of the projected probabilities. We then recorded the ratio of times that the ground truth inferences fell within the projected confidence intervals. This ratio $r(\gamma)$, when compared to the associated significance level $\gamma$, gives the DeCBoD for each particular value of $\gamma$. We allowed $\gamma$ to range from 0 to 1 in increments of 0.01. The DeCBoD metric was chosen so that our results could be directly compared to the prior literature, namely \cite{kaplan}, and a more thorough description of the DeCBoD metric is contained in \cite{kaplan-ivanovska}.

With this basic framework, we performed two main experiments, which we call ``Experiment A'' and ``Experiment B.'' These experiments were designed to demonstrate some of the key differences between the three learning algorithms. In particular, the BMM method assumes statistical independence between separate families of parameters. We expect this assumption to cause problems when the training data are incomplete, since families of conditional probabilities become statistically dependent in those cases. This principle is exacerbated when the variables observed share a common ancestor for which the observations are limited. Experiment B, in which the leaf nodes make up the vast majority of observations, produces such statistical dependencies. Experiment A, then, is our baseline experiment which allows us to compare general algorithmic performance for learning the posterior distributional covariance, as well as the computational efficiency, given purely random observations.

In Experiment A, for each of the $N=1000$ ground truth BNs, we generated $T=120$ observations of the BN state variables, then retained a fraction of those observations as training data. The retention fractions were in the range $f\in\{0.1, 0.2, \ldots, 0.9\}$. We also varied the structure of the BN between experiments. See Figure \ref{fig:BNarchitectures} for the particular structures of interest, which included both a 3-node chain and a representative 9-node DAG. We used three-valued categorical variables in each of the networks. The purpose of Experiment A was to quantify the mean absolute DeCBoD, i.e., 
\begin{equation}
   \frac{1}{101} \sum_{\gamma=0:.01:1} |r(\gamma)-\gamma|,
\end{equation} for each of the three learning algorithms discussed in Section \ref{meth}, under the constraint of random partial training data. The DeCBoD was measured after incorporation of state-of-the-art second-order inferencing in \cite{cerutti} for each of the queried probabilities, i.e. for probabilities of form $p(x_k|\bm{e}), \forall k\text{ such that } X_k\notin \bm{E}$. Note that during the inferencing step, the evidence variables, $\bm{E}$, and their respective values, were randomly chosen.

In Experiment B, for each of the $N=1000$ ground truth BNs, we seeded the learning algorithms with a set of 20 complete observations of the BN, followed by 100 partial training examples of a pre-specified form, to give $T=120$ total observations of each network. In particular, for the 100 partial training examples, we chose to only observe the leaf nodes of the BN. Our motivation for observing only the leaf nodes was to introduce strong statistical dependencies between families of the BN parameters. Leaf node observations are a straightforward way to effectively induce statistical dependencies, but any set of observations where the observed variables share a common ancestor would also suffice, including only a partial observation of some of the leaf nodes. For this experiment, we observed all of the leaf nodes for simplicity. It is straightforward to repeat the experiment with other sets of observed variables. Specifically, we again used the 9-node DAG structure from Figure \ref{fig:BNarchitectures}(b), and the 100 partial observations were of the leaf-node variables $X_6, X_7,$ and $X_8$. 

For some intuition on why the statistical independence assumption is violated when leaf nodes are observed, consider the 9-node DAG in Figure \ref{fig:BNarchitectures}(b). Specifically, consider the tree formed by $X_5,X_7,$ and $X_8$. By observing $X_7$ and $X_8$ without observing $X_5$, the parameters governing the links $X_5 \rightarrow X_7$ and $X_5 \rightarrow X_8$ become statistically dependent. This is because any estimates of parameters $P(X_7=k|X_5=1), \forall k\in \mathbb{X}_7$, when combined with the actual observed value of $X_7$, tells us something about the value of $X_5$. In turn, this implicit information affects any reasonable estimates of parameters $P(X_8=k|X_5=1), \forall k\in \mathbb{X}_8.$


The experiments were run using MATLAB R2020a, on a PC with 16GB of RAM and an Intel\textregistered i7-4790K CPU with 64-bit Windows 10.

\subsection{Results and Discussion: Experiment A}
\begin{figure}[t]
\centering
\subfloat[]{\includegraphics[width=1.75in]{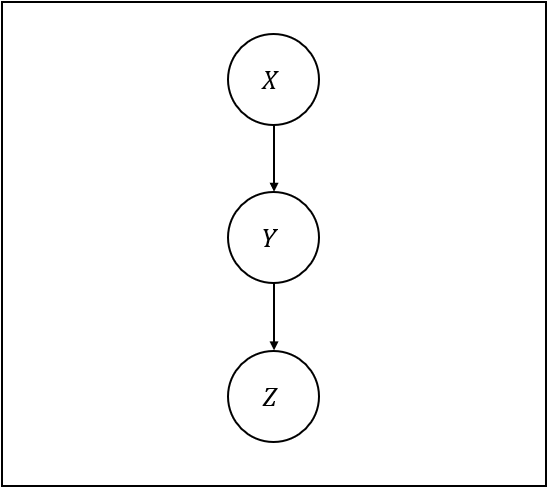}}
\subfloat[]{\includegraphics[width=1.75in]{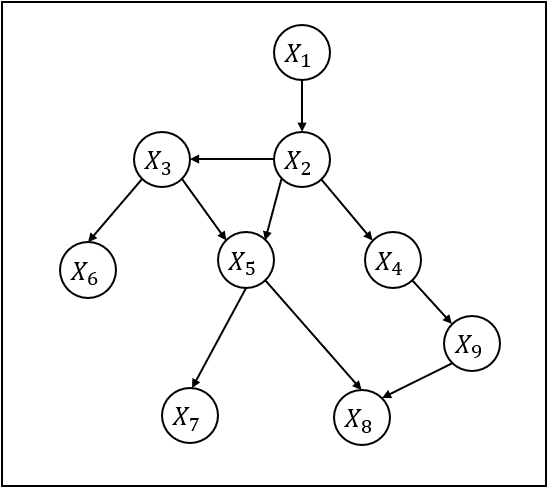}} 
\caption{Examples of (a) 3-node chain and (b) 9-node DAG.} 
\label{fig:BNarchitectures} 
\end{figure}

\begin{figure}
    \centering
    \includegraphics[scale=0.26]{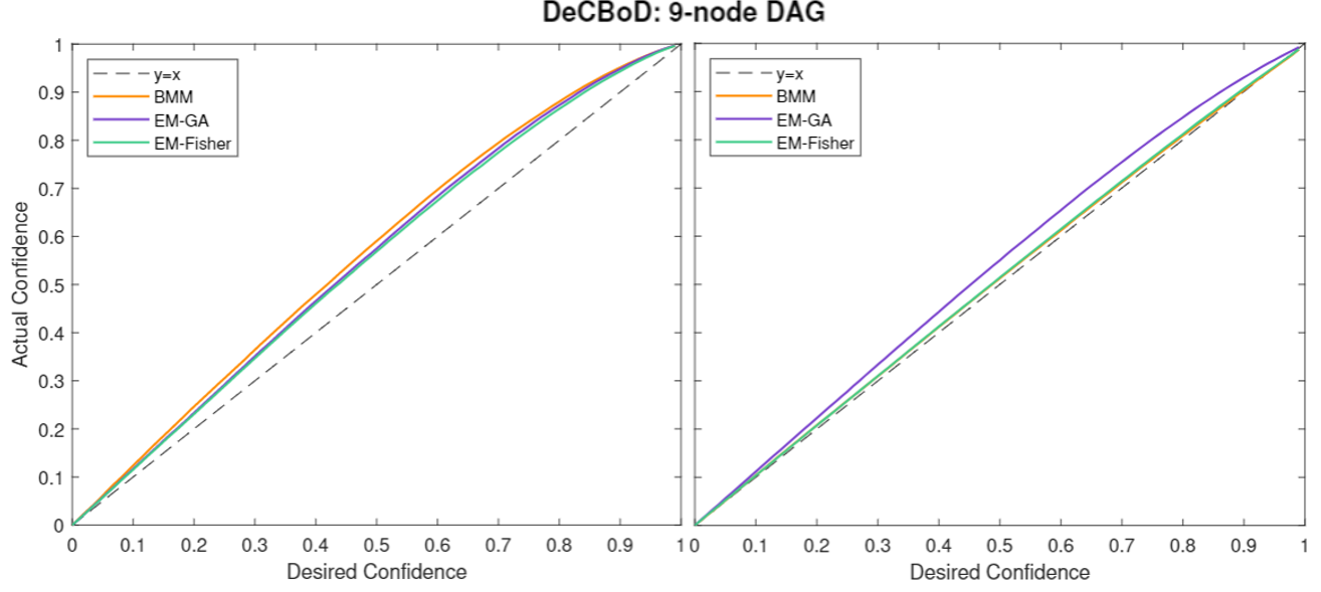}
    \caption{Experiment A: DeCBoD for 9-node DAG with randomly observed variables and $f=0.1$ (left) and $f=0.5$ (right).}
    \label{fig:9noderandom}
\end{figure}

Experiment A resulted in DeCBoD curves similar to the ones shown in Figure \ref{fig:9noderandom}. In the left sub-figure, the fractional observation rate is $f=0.1$, while in right sub-figure, the fractional observation rate increases to $f=0.5$. Both the left and right sub-figures correspond to the particular choice of 9-node DAG shown in Figure \ref{fig:BNarchitectures}(b). From the DeCBoD curves, we observe dramatic convergence to the diagonal line when the fractional observation rate increases. In particular, EM-Fisher and BMM reproduce the posterior distributions of the conditional probabilities well, resulting in the desired confidence levels nearly matching the actual confidence levels. These DeCBoD curves correspond to the entries in the bottom half of Table \ref{tbl:decbodtbl}, under $f=0.1$ and $f=0.5$. The table contains the complete mean absolute DeCBoD values for the 3-node chain and 9-node DAG, for all values of $f$ from $0.1$ to $0.9$.

From Table \ref{tbl:decbodtbl}, for both the 3-node chain and 9-node DAG cases, the mean absolute DeCBoD decreases rapidly as the fractional observation rate, $f$, increases, for BMM and EM-Fisher. On the other hand, the mean absolute DeCBoD for EM-GA also decreases somewhat, but not nearly as consistently or dramatically as in the other two methods. EM-GA struggles here, primarily because of the relatively low number of training examples, with $T\approx 100$. For any particular parameter, the number of training examples which correspond to that parameter is only a fraction of $T$. In essence, $T$ is not large enough here for the Hessian to match the Fisher information matrix. On the other hand, EM-Fisher outperforms EM-GA at every value of $f$, which provides empirical evidence that justifies the Bernstein von-Mises theorem, which says that the covariance matrix should be well-approximated by the inverse Fisher information matrix. On the other hand, BMM appears to outperform both methods when more of the data are complete, i.e. for $f> 0.5$.

\begin{table}[h]
    \centering
      \caption{Experiment A: mean absolute DeCBoD for each method for a 3-node and 9-node BN, with varying fractional observation rates of the network variables and $T=120$.}
   \label{tbl:decbodtbl}
   \includegraphics[scale=0.6]{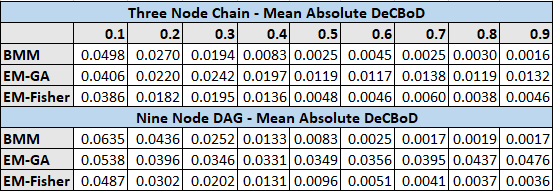}
 
\end{table}

\begin{table}[h]
    \centering
    \caption{Experiment A: mean wall-clock times for each method for a 3-node and 9-node BN, with varying fractional observation rates of the network variables and $T=120$.}
    \includegraphics[scale=0.6]{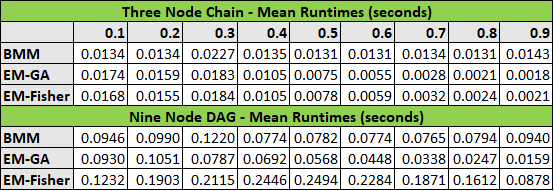}
    \label{tbl:wallclktbl}
\end{table}

In Experiment A, we also profiled the learning methods to gauge computational efficiency and scalability. In particular, we wanted to know which method would be most impacted by a transition to larger networks with more nodes. It turns out that EM-Fisher takes the most wall-clock time to complete for networks of moderate size, such as the 9-node DAG that we have been using as our exemplar in this paper. Table \ref{tbl:wallclktbl} shows that EM-GA runs increasing quickly as $f$ increases. A similar trend can be seen for EM-Fisher for the 3-node case. This is due to EM requiring fewer iterations to converge when the data are more complete. Table \ref{tbl:wallclktbl} also shows that, as the network size increases from 3 to 9 nodes, BMM scales best, while the EM methods do not scale as well. This can be at least partially attributed to the matrix inversion steps of the EM methods, shown in eqns. (\ref{eqn:EM-GA_inversion}) and (\ref{eqn:EM-Fisher_inversion}). Also, EM-Fisher scales the worst and runs slowest for $f$ near $0.5$ for the 9-node network. We believe this is due to the number of possible variable values in the inner sum in eqn. (\ref{eqn:EM-Fisher}).

\subsection{Results and Discussion: Experiment B}
\begin{figure}
    \centering
    \includegraphics[scale=0.14]{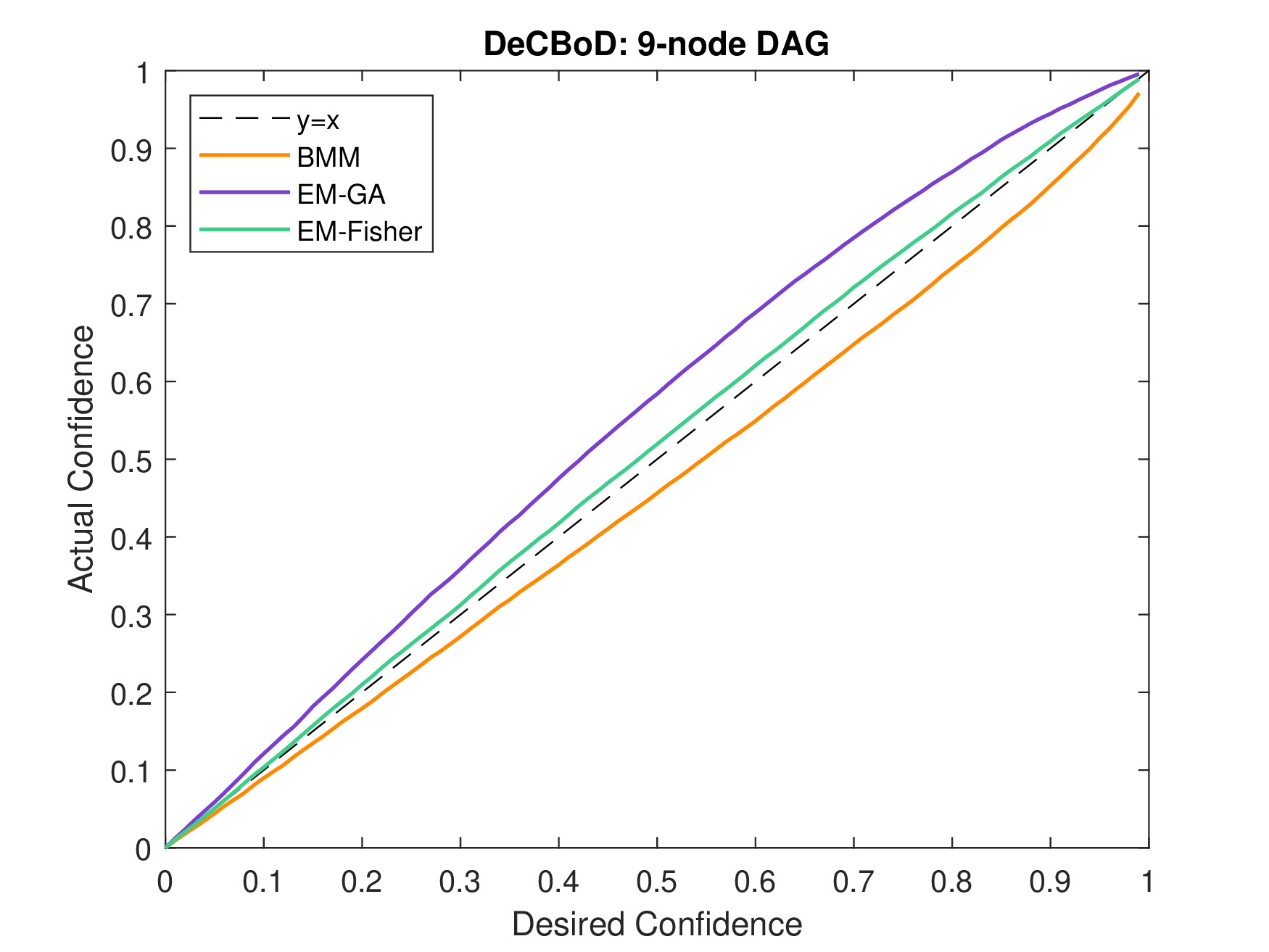}
    \caption{Experiment B: DeCBoD for 9-node DAG with leaf nodes observed.}
    \label{fig:decbod_9nodedag_expB}
\end{figure}

Figure \ref{fig:decbod_9nodedag_expB} is the resulting DeCBoD plot corresponding to Experiment B, in which we only observed the leaf nodes of the 9-node DAG. Of interest, BMM underestimates the desired confidence level, which is why the BMM DeCBoD curve (orange) lies below the diagonal. This means that BMM provides a confidence bound that is too tight, i.e. BMM is underestimating the uncertainty. This matches our expectations. Recall that BMM assumes statistical independence of the parameter families, thereby resulting in a block-diagonal approximation  of the covariance matrix. However, in Experiment B, only the leaf nodes of the 9-node DAG were observed, inducing strong dependencies between the parameters. In fact, parameter independence is violated in the incomplete data case in general. In Experiment A, the randomness of the observations allowed for the conditional probabilities to be well-approximated, regardless. In essence, BMM ignores off-diagonal covariance elements which would increase the uncertainty in our parameter estimates. EM-GA and EM-Fisher are immune from these effects since the full covariance matrix approximations are generated, using the Hessian and Fisher information matrices, respectively.

\vspace{0.1in}

\section{Conclusion}\label{conc}
In this work, we provided the first evaluations of algorithms capturing both the mean and covariance of the uncertain conditional probabilities for later second-order inferencing. We considered techniques based upon method of moments, Gaussian approximation and the Fisher information matrix, that we extensively tested in terms of (1) accuracy of the computation of the covariance matrix, which lies at the heart of uncertain Bayesian networks, through consistency of the desired and empirically derived strength of confidence bounds; and (2) their scalability. 


To our knowledge, no prior works have applied these methods to the setting of incomplete training data over networks of more than two nodes. Kaplan et. al. introduced a simple two-node binary-valued example in \cite{kaplan}. In this work, we have extended that simple example further, demonstrating the applicability of additional second-order learning methods to Bayesian networks of theoretically arbitrary size.

From the results contained in this paper, we can conclude that BMM outscales EM-Fisher and rapidly becomes the method of choice as the number of variables and parameters increases. Furthermore, with the exception of certain pathological cases, BMM also provides relatively accurate estimations of the parameter uncertainty, as shown by the DeCBoD curves and mean absolute DeCBoD values generated by our simulations. EM-GA provides worse performance on the limited data cases for the network structures that we focused on in this work. However, by the law of large numbers, we do know that Hessian computation should asymptotically match the Fisher information,
meaning that we would expect EM-GA to improve in cases where we have more training data available than 120 samples. Unfortunately, high data availability may not be realistic for many practical use cases.

There are several avenues we are exploring for future work. First, we aim to further improve the scalability of the methods described here. Second, we have envisaged a comprehensive comparison with Monte Carlo approaches to assess the effect of our approximations using Dirichlet distributions. Third, it should be mentioned that the current experiments have some shortcomings. For example, Experiment B utilizes a small seed set of complete observations prior to the leaf-node only observations. We would like to expand on these results by determining how many seed observations are needed, on average, before the Fisher information matrix (FIM) is at least approximately invertible. We seek to understand the relationship between quantity of training data and the FIM condition number. We would also like to test partial observations of variables which are not leaf nodes. In Experiment B, we restricted our training observations to leaf nodes in order to induce statistical correlations between families of variables. It would benefit our work to also quantify the effects of using other sets of partial data for training, including non-leaf nodes.

\bibliographystyle{IEEEtran}
\bibliography{ref}

\begin{thebibliography}{10}
\providecommand{\url}[1]{#1}
\csname url@samestyle\endcsname
\providecommand{\newblock}{\relax}
\providecommand{\bibinfo}[2]{#2}
\providecommand{\BIBentrySTDinterwordspacing}{\spaceskip=0pt\relax}
\providecommand{\BIBentryALTinterwordstretchfactor}{4}
\providecommand{\BIBentryALTinterwordspacing}{\spaceskip=\fontdimen2\font plus
\BIBentryALTinterwordstretchfactor\fontdimen3\font minus
  \fontdimen4\font\relax}
\providecommand{\BIBforeignlanguage}[2]{{%
\expandafter\ifx\csname l@#1\endcsname\relax
\typeout{** WARNING: IEEEtran.bst: No hyphenation pattern has been}%
\typeout{** loaded for the language `#1'. Using the pattern for}%
\typeout{** the default language instead.}%
\else
\language=\csname l@#1\endcsname
\fi
#2}}
\providecommand{\BIBdecl}{\relax}
\BIBdecl

\bibitem{VanAllen}
T.~Van~Allen, R.~Greiner, and P.~Hooper, ``Bayesian error-bars for belief net
  inference,'' in \emph{2001 17th Conference in Uncertainty in Artificial
  Intelligence}.\hskip 1em plus 0.5em minus 0.4em\relax AUAI, 2001, pp.
  522--529.

\bibitem{shenoy}
P.~P. Shenoy, ``A valuation-based language for expert systems,''
  \emph{International Journal of Approximate Reasoning}, vol.~3, no.~5, pp.
  383--411, 1989.

\bibitem{zaffalon}
E.~Fagiuoli and M.~Zaffalon, ``2u: an exact interval propagation algorithm for
  polytrees with binary variables,'' \emph{Artificial Intelligence}, vol. 106,
  no.~1, pp. 77--107, 1998.

\bibitem{kaplan-ivanovska}
L.~Kaplan and M.~Ivanovska, ``Efficient belief propagation in second-order
  bayesian networks for singly-connected graphs,'' \emph{International Journal
  of Approximate Reasoning}, vol.~93, pp. 132--152, 2018.

\bibitem{rohmer}
J.~Rohmer, ``Uncertainties in conditional probability tables of discrete
  {Bayesian} belief networks: A comprehensive review,'' \emph{Engineering
  Applications of Artificial Intelligence}, vol.~88, p. 103384, 2020.

\bibitem{cerutti}
F.~Cerutti, L.~M. Kaplan, A.~Kimmig, and M.~Sensoy, ``Handling epistemic and
  aleatory uncertainties in probabilistic circuits,'' \emph{arXiv preprint
  arXiv:2102.10865}, 2021.

\bibitem{kaplan}
L.~Kaplan, F.~Cerutti, M.~{\c{S}}ensoy, and K.~V. Mishra, ``Second-order
  learning and inference using incomplete data for uncertain {Bayesian}
  networks: A two node example,'' in \emph{2020 IEEE 23rd International
  Conference on Information Fusion (FUSION)}.\hskip 1em plus 0.5em minus
  0.4em\relax IEEE, 2020, pp. 1--8.

\bibitem{nielsen}
F.~V. Jensen and T.~D. Nielsen, \emph{Bayesian networks and decision
  graphs}.\hskip 1em plus 0.5em minus 0.4em\relax Springer, 2001, vol.~2.

\bibitem{Zhao}
H.~Zhao, M.~Melibari, and P.~Poupart, ``On the relationship between sum-product
  networks and bayesian networks,'' in \emph{Proceedings of the 32nd
  International Conference on Machine Learning}, ser. Proceedings of Machine
  Learning Research, vol.~37.\hskip 1em plus 0.5em minus 0.4em\relax PMLR,
  07--09 Jul 2015, pp. 116--124.

\bibitem{darwiche}
A.~Darwiche, ``A differential approach to inference in bayesian networks,''
  \emph{Journal of the ACM (JACM)}, vol.~50, no.~3, pp. 280--305, 2003.

\bibitem{rashwan}
A.~Rashwan, H.~Zhao, and P.~Poupart, ``Online and distributed bayesian moment
  matching for parameter learning in sum-product networks,'' in
  \emph{Artificial Intelligence and Statistics}.\hskip 1em plus 0.5em minus
  0.4em\relax PMLR, 2016, pp. 1469--1477.

\end{thebibliography}




\end{document}